\documentclass[letter]{ieice}
\usepackage[pdftex]{graphicx,xcolor}
\usepackage[fleqn]{amsmath}
\usepackage{newtxtext}
\usepackage[varg]{newtxmath}
\usepackage{mathtools}

\usepackage{algorithm,algorithmic}

\usepackage{booktabs}
\usepackage{multirow}
\usepackage{diagbox}
\usepackage{url}

\field{D}
\title{MNIST-MIX: A Multi-language Handwritten Digit Recognition Dataset}
\authorlist{%
\authorentry[jiangweiwei@mail.tsinghua.edu.cn]{Weiwei Jiang}{m}{beijing}\MembershipNumber{1}
}
\affiliate[beijing]{The author is with Department of Electronic Engineering, Tsinghua University, Beijing 100084, China.
}
\received{2015}{1}{1}
\revised{2015}{1}{1}

\begin{document}
\maketitle
\begin{summary}
In this letter, we contribute a multi-language handwritten digit recognition dataset named MNIST-MIX, which is the largest dataset of the same type in terms of both languages and data samples. With the same data format with MNIST, MNIST-MIX can be seamlessly applied in existing studies for handwritten digit recognition. By introducing digits from 10 different languages, MNIST-MIX becomes a more challenging dataset and its imbalanced classification requires a better design of models. We also present the results of applying a LeNet model which is pre-trained on MNIST as the baseline.
\end{summary}
\begin{keywords}
Deep Learning, Handwritten Digit Recognition, Convolutional Neural Network 
\end{keywords}

\section{Introduction}
%\vfill\pagebreak
handwritten digit recognition is a classical but important problem in computer vision, which has a wide application in different areas and can be embedded into larger systems. As a popular benchmark dataset, MNIST~\cite{1} has been widely used for benchmarking different recognition models. Nowadays, with the development of deep learning represented by convolutional neural networks, MNIST becomes too easy for modern deep learning models. Even a CNN model with only three layers can achieve an accuracy better than 99\% on MNIST~\cite{16}, which indicates that MNIST is not enough for the performance evaluation of more sophisticated models.

One direction of making MNIST more challenging is to increase the size of MNIST, by incorporating more images. EMNIST~\cite{2}, as the name indicated, is an extension of MNIST to handwritten letters. EMNIST also provides a larger sample size for digits. We would combine EMNIST, instead of MNIST, into MNIST-MIX, as MNIST is a subset of EMNIST. Another direction is to collect digits from real-world scenes. Instead of collected from designed experiments, the Street View House Numbers (SVHN) dataset~\cite{12} is obtained from house numbers in Google Street View images and aims to build up as a significantly harder and unsolved real-world problem, \textit{i.e.}, recognizing digits and numbers in natural scene images. The third direction is to extend handwritten characters to other objects. Fashion-MNIST~\cite{13} is a dataset of Zalando's article images, which uses the same data format and sample sizes as MNIST and has been widely used for benchmarking different machine learning and deep learning models.

As another direction of extension for MNIST, the work of developing a multi-language version is less discussed in literature. The recognition of mixed handwritten numerals of three Indian scripts Devanagari, Bangla and English is considered in~\cite{14} and handwritten characters from multi-language document images, which may contain various types of characters, e.g. Chinese, English, Japanese or their mixture, are extracted in~\cite{15}.

In this letter, we contribute MNIST-MIX, the largest multi-language handwritten digit recognition dataset as the author knows. We also contribute a pre-trained LeNet model that achieves an accuracy of 90.22\% on MNIST-MIX. For further studies, MNIST-MIX and the code to reproduce the results in this letter are publicly available ~\footnote{Available: \url{https://github.com/jwwthu/MNIST-MIX}}.

\section{Data Collection\&Processing}
In this letter, we combine 13 different datasets for 10 languages (\textit{i.e.}, Arabic, Bangla, Devanagari, English, Farsi, Kannada, Swedish, Telugu, Tibetan, and Urdu) into MNIST-MIX. During the combination, we perform the following processing steps to make sure that data samples from different sources share the same data format with MNIST, \textit{i.e.}, each input image has a resolution of $28 \times 28$, in gray-scale format, and corresponds to a number between 0 and 9:

(1) RGB to gray-scale conversion: Some of the images are in color space, \textit{i.e.}, red-green-blue space. To convert them into a gray-scale format, we use the following formulation: $L = R \times 299/1000 + G \times 587/1000+ B \times 114/1000$, and each pixel is represented by an integer in the range of $[0, 255]$, where 0 represents black and 255 represents white.

(2) Resizing: Since not all datasets have the same data format with MNIST, we resize some of them to achieve uniformity in format. We implement this function with the resize function provided by OpenCV, with a bilinear interpolation.

(3) Training/Testing split: Some of the datasets have no explicit split of training/testing split. For these datasets, we split the whole dataset with a training/testing ratio of 80\%:20\%. During the split process, the percentage of samples for each class is preserved.

Specifically, MNIST-MIX comes from a combination of the following open datasets:

(1) ARDIS~\cite{4}: Arkiv Digital Sweden (ARDIS) is extracted from 15,000 Swedish church records, and it provides image samples both in color and in gray, which saves us the operation of RGB to gray-scale conversion. The part in the MNIST format that we use contains 6,600 training and 1,000 testing images.

(2) BanglaLekha-Isolated~\cite{7}: Bangla is mainly used in Bangladesh and West Bengal. BanglaLekha-Isolated contains Bangla handwritten numerals, basic characters and compound characters, and was collected from Bangladesh. Another feature of BanglaLekha-Isolated is that it contains more metrics of the writers, \textit{e.g.}, gender, age, districts. Each image sample has a different size and we unify them into the MNIST format.

(3) CMATERdb database~\footnote{\url{https://code.google.com/archive/p/cmaterdb/}}, from which we use four separate datasets: CMATERdb 3.1.1 (handwritten Bangla numeral database), CMATERdb 3.2.1 (handwritten Devanagari numeral database), CMATERdb 3.3.1 (handwritten Arabic numeral database), CMATERdb 3.4.1 (handwritten Telugu numeral database). CMATERdb is a open access database created by the Cmater Research Laboratory for Training Education and Research in Jadavpur University, India. Bangla, Devanagari, Arabic and Telugu are all languages used in some areas of India. We combine 6,000 Bangla numerals, 3,000 Devanagari numerals, 3,000 Arabic numerals, and 3,000 Telugu numerals into MNIST-MIX with a manual training/testing split.

(4) EMNIST~\cite{2}: As a natural extension of MNIST, EMNIST is derived from the NIST Special Database 19 and converted to a 28$\times$28 pixel image format and dataset structure that directly matches MNIST. EMNIST contains both letters and digits, but we only use the part of digits in MNIST-MIX.

(5) FARSI~\cite{5}: Farsi is a Western Iranian language that is used officially within Iran, Afghanistan and Tajikistan. This dataset is extracted from about 12,000 registration forms of two types, filled by B.Sc. and senior high school students.

(6) ISI Bangla~\cite{6}: It is another dataset for Bangla, which is collected from 465 mail pieces and 268 job application forms. 

(7) Kannada-mnist~\cite{8}: Kannada-MNIST is a handwritten digit dataset for the Kannada language, which is used predominantly by people of Karnataka in southwestern India. Kannada-MNIST consists of 60,000 training samples and 10,000 testing samples, with the same data format with MNIST. Kannada-MNIST is accompanied by Dig-MNIST, which is an out-of-domain test dataset, which is collected by volunteers who are not native speakers of Kannada and contains 10,240 testing samples. We combine the original testing set of Kannada-MNIST and Dig-MNIST as the new testing set with a size of 20,240.

(8) MADBase~\footnote{\url{http://datacenter.aucegypt.edu/shazeem/}}: MADBase is created from a larger ADBase, which is collected from writings by 700 participants, and shares the same format with MNIST for the Arabic digit recognition problem.

(9) TibetanMNIST~\footnote{\url{https://www.kesci.com/home/dataset/5bfe734a954d6e0010683839}}: It is the first open dataset of Tibetan handwritten numerals, which is used by 6.5 million Tibetan people mainly living in southwest China. We transform the RGB format to the gray-scale format and resize the original data from $32 \times 32$ to $28 \times 28$, to keep consistent with the desired size. The dataset is manually divided into a training set with 14,214 samples and a testing set with 3,554 samples.

(10) Urdu~\cite{11}: Urdu language is used by more than 60 million people and this dataset is collected from more than 900 individuals. We only use the part of digits.

\begin{figure}[!htb]
    \centering
    \includegraphics[width=0.45\textwidth]{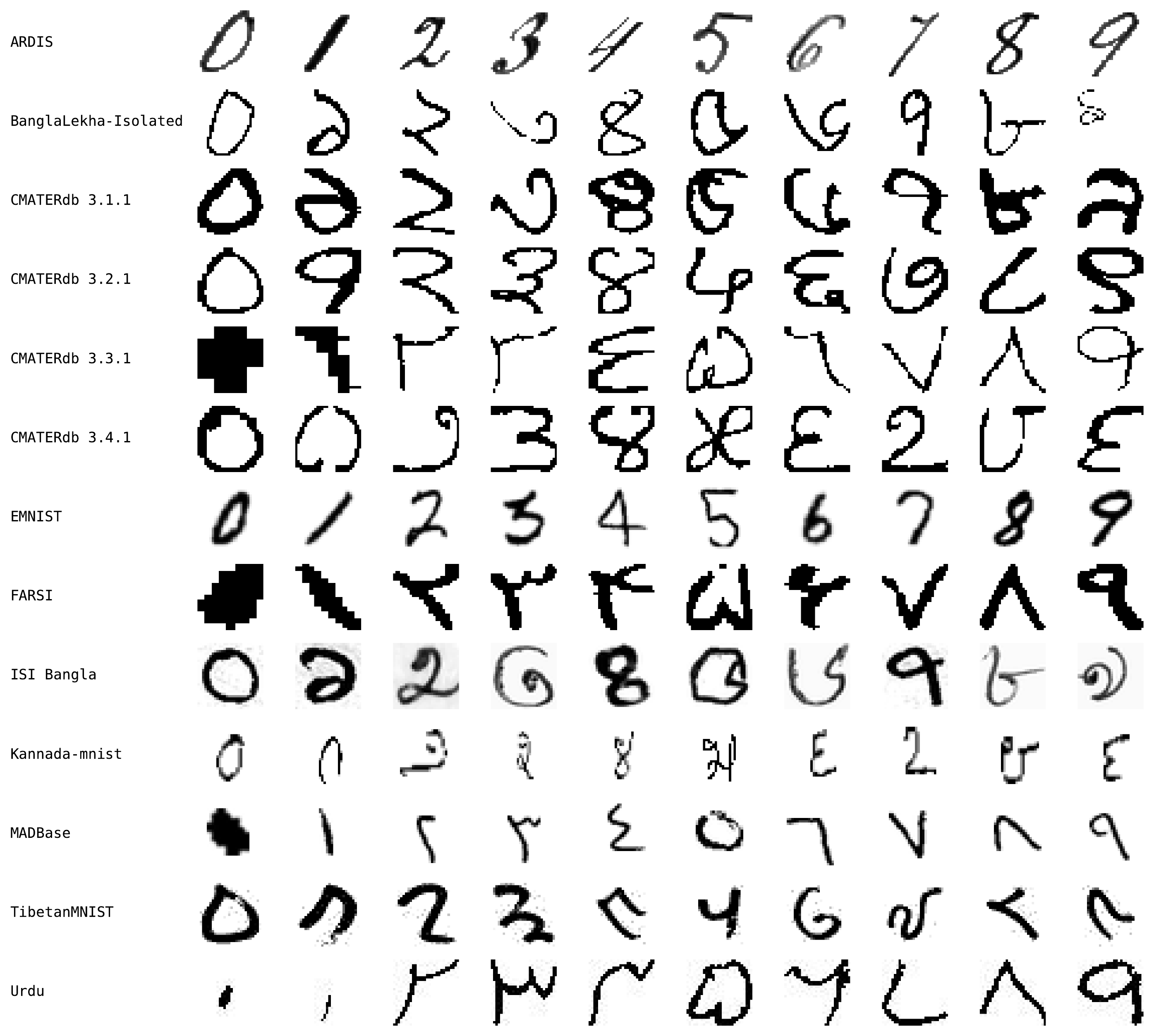}
    \caption{Examples from different handwritten numeral datasets.}
    \label{fig:datasets}
\end{figure}

\begin{table}[!htb]
    \centering
    \caption{Dataset list with training and testing sizes.}
    \begin{tabular}{|l|l|l|l|}
        \hline
        Dataset Name & Language & Training Size &  Testing Size \\
        \hline
        ARDIS & Swedish & 6,600 & 1,000 \\
        BanglaLekha-Isolated & Bangla & 15,798 & 3,950\\
        CMATERdb 3.1.1 & Bangla & 4,800 & 1,200 \\
        CMATERdb 3.2.1 & Devanagari & 2,400 & 600 \\
        CMATERdb 3.3.1 & Arabic & 2,400 & 600 \\
        CMATERdb 3.4.1 & Telugu & 2,400 & 600 \\
        EMNIST & English & 240,000 & 40,000 \\
        FARSI & Farsi & 60,000 & 20,000 \\
        ISI Bangla & Bangla & 19,392 & 4,000 \\
        Kannada-mnist & Kannada & 60,000 & 20,240 \\
        MADBase & Arabic & 60,000 & 10,000 \\
        TibetanMNIST & Tibetan & 14,214 & 3,554 \\
        Urdu & Urdu & 6,606 & 1,414 \\
        \hline
    \end{tabular}
    \label{tab:datasets}
\end{table}

We show the digits from different datasets after pre-processing in Figure~\ref{fig:datasets}. We also show the training and testing sizes in Table~\ref{tab:datasets}. As shown in Table~\ref{tab:datasets}, MNIST-MIX is a highly unbalanced dataset and the data samples for different languages could vary a lot, which becomes a challenge for designing the recognition models.

\section{Experiment}
As a pioneer deep learning model, LeNet was proposed with the MNIST dataset. It consists standard layer structures, \textit{e.g.}, fully connected layer, convolutional layer, pooling layer, etc. LeNet comprises 7 layers without counting the input layer, in which two sets of the combination of convolutional layer and sub-sampling layer are used to extract features. An output layer with softmax as the activation function is used to predict the probabilities of the sample class.

We firstly train the LeNet model on the original MNIST dataset, which achieves a test accuracy of 99.26\%. Then we freeze the network connections and only fine-tune the parameters of the output layer on each individual dataset. For MNIST-MIX, we also change the output layer's size to 100, as MNIST-MIX contains 100 classes. 

We implement the LeNet model with the Python packages Keras and Tensorflow. For training the model, we use Adam with a learning rate 0.001 as the optimizer and categorical cross entropy as the loss function. Early stopping is used for prevention of overfitting. The experiments are conducted on a desktop computer that is equipped with a Windows 10 operating system, 16GB random-access memory (RAM), Intel core i5-9600K central processing unit (CPU) and a graphical accelerated processing (GPU) of GeForce RTX 2070 with 8GB RAM.

We use three metrics for evaluation, i.e., accuracy, weighted F1 score and balanced accuracy. Accuracy is defined as the percentage of test samples that are correctly classified, where an accuracy reaches its best value at 100\% and worst value at 0\%. F1 score is the average of the precision and recall, where an F1 score reaches its best value at 1 and worst score at 0. The weighted F1 is calculated for each class and averaged with support (the number of true samples for each class) as weight. Balanced accuracy is defined as the average of recall obtained on each class, where a balanced accuracy also reaches its best value at 100\% and worst value at 0\%.

The result of applying the pre-trained LeNet model on separate datasets and MNIST-MIX is shown in Table~\ref{tab:experiments}. The LeNet model achieves the best for EMNIST, which is reasonable because EMNIST and MNIST share the same source of data samples. For MNIST-MIX, the LeNet model can achieve an accuracy of 90.22\%, but the balanced accuracy is only 66.61\%, which indicates the performance degrades a lot when considering the highly imbalanced class problem and leaves a large space for further improvement.

\begin{table}[!htb]
    \centering
    \caption{The results of applying the pre-trained LeNet model.}
    \begin{tabular}{|l|l|l|l|}
        \hline
        Dataset Name & Accuracy & Weighted F1 &  Balanced Accuracy \\
        \hline
        ARDIS & 0.9820 & 0.9820 & 0.9820 \\
        BanglaLekha-Isolated & 0.9486 & 0.9484 & 0.9486 \\
        CMATERdb 3.1.1 & 0.9550 & 0.9550 & 0.9550 \\
        CMATERdb 3.2.1 & 0.9700 & 0.9700 & 0.9700 \\
        CMATERdb 3.3.1 & 0.9517 & 0.9514 & 0.9517 \\
        CMATERdb 3.4.1 & 0.9783 & 0.9783 & 0.9783 \\
        EMNIST & 0.9950 & 0.9950 & 0.9950 \\
        FARSI & 0.9818 & 0.9818 & 0.9818 \\
        ISI Bangla & 0.9705 & 0.9704 & 0.9705 \\
        Kannada-mnist & 0.8570 & 0.8562 & 0.8570 \\
        MADBase & 0.9893 & 0.9893 & 0.9893 \\
        TibetanMNIST & 0.9828 & 0.9828 & 0.9831 \\
        Urdu & 0.9731 & 0.9731 & 0.9727 \\
        \hline
        MNIST-MIX & \textbf{0.9022} & \textbf{0.8910} & \textbf{0.6661} \\
        \hline
    \end{tabular}
    \label{tab:experiments}
\end{table}

\section{Conclusion}
Based on the thorough collection of available handwritten digit datasets, we build and release MNIST-MIX, a multi-language handwritten digit recognition dataset, as a drop-in replacement for the original MNIST dataset with the same image format. We also present a pre-trained LeNet model on MNIST as the baseline for the task of multi-language handwritten digit recognition.

% \section*{Acknowledgments}
% The author(s) received no specific funding for this work.

\bibliographystyle{ieicetr}% bib style
\bibliography{main}% your bib database
%\begin{thebibliography}{99}% more than 9 --> 99 / less than 10 --> 9
%\bibitem{}
%\end{thebibliography}

%\profile{}{}
%\profile*{}{}% without picture of author's face

\end{document}